\documentclass[10pt,letterpaper]{article}

\usepackage{graphicx}
\usepackage{multirow}

\usepackage{cogsci}

\usepackage{pslatex}
\usepackage{apacite}
\usepackage{float} 

\title{A Neural Network Model of Spatial and Feature-Based Attention}
 
\author{{\large \bf Ruoyang Hu (rhu13@ur.rochester.edu)} \\
  Department of Brain and Cognitive Sciences, University of Rochester \\
  Rochester, NY 14627 USA
  \AND {\large \bf Robert A. Jacobs (robbie@bcs.rochester.edu)} \\
  Department of Brain and Cognitive Sciences, University of Rochester \\
  Rochester, NY 14627 USA}

\begin{document}

\maketitle

\begin{abstract}
Visual attention is a mechanism closely intertwined with vision and memory. Top-down information influences visual processing through attention. We designed a neural network model inspired by aspects of human visual attention. This model consists of two networks: one serves as a basic processor performing a simple task, while the other processes contextual information and guides the first network through attention to adapt to more complex tasks. After training the model and visualizing the learned attention response, we discovered that the model’s emergent attention patterns corresponded to spatial and feature-based attention. This similarity between human visual attention and attention in computer vision suggests a promising direction for studying human cognition using neural network models.

\end{abstract}

\section{Introduction}

We are constantly exposed to an overwhelming amount of perceptual input from the external world. Even when the eyes remain fixed and the retinal image does not change, the sheer volume of visual information can still be difficult to process. For instance, in the classic puzzle book series \textit{Where’s Waldo}, readers are tasked with locating a character named Waldo amidst a highly detailed and cluttered background. Humans cannot process all visual information simultaneously, partly due to the biological energy constraints of neuronal activity \cite{lennie2003cost}. Attention acts as a selective mechanism, allowing us to focus on stimuli most relevant to the task at hand. This selective process, when the eyes are fixed, is known as covert visual attention.

Covert attention can be further divided into two processes: exogenous and endogenous attention. Exogenous attention is a bottom-up mechanism driven by salient visual features, allowing us to detect changes in the visual field and respond quickly—critical for survival. In contrast, endogenous attention is a top-down process where cognitive factors such as prior knowledge, expectations, and goals influence perception. Higher-level cognitive processes can guide attention in a top-down manner, prioritizing specific information.

Top-down visual selection has commonly shown two different kinds: spatial attention and feature-based attention. Spatial attention is often metaphorically described as a “spotlight” that enhances resolution in the attended region \cite{tsal1995towards, intriligator2001spatial}. Feature-based attention, on the other hand, focuses on specific visual features like color, orientation, or motion, regardless of their spatial locations.

Feature dimensions or locations are selected through the top-down control. Contextual information that hints at what is relevant to the task or the goal is top-down information that can guide attention. For example, if you are searching for a book you remember placing on the upper shelf, your attention will naturally shift to that area. Additionally, knowing the book has a red cover will direct your attention to the red books on the shelf.

While attention mechanisms have been widely applied in computer vision models, and many computational models aim to study visual attention at the neuronal level, discrepancies remain between attention as implemented in computer vision and how it functions in humans. To address this, we designed a neural network system that simulates the top-down control of human visual attention. We believe that exploring connections between these areas can offer valuable insights.

\citeA{pollack1987cascaded} proposed a neural network system well-suited for simulating the top-down selection of visual attention. The system has two networks: a context network processes top-down contextual information and helps to guide the attention of a function network. Inspired by Pollack’s paper, we designed a similar system to demonstrate top-down control, proposing that it should perform similarly to human attention. If this hypothesis holds true, we can conclude that attention can be learned and that the system’s performance is guided by top-down context. Conversely, if the results do not align, it may indicate inherent differences between human and computer performance or flaws in the proposed model.

The proposed attention model simulates how top-down contextual information is applied to guide selection. This process reflects the top-down control present in human visual attention and represents a plausible model that suggests a promising direction for comparing human and computer visual attention, facilitating a more collaborative investigation of attention mechanisms in both domains.

\section{Related Works}
\subsection{Transformer}
The Transformer is a widely used neural network architecture built around an attention mechanism. It was first introduced in the paper ``Attention is All You Need'' \cite{vaswani2017attention}. A standard Transformer includes a self-attention mechanism, an encoder-decoder structure, and positional encoding. The self-attention mechanism determines the importance of each word in an input sentence relative to all other words, rather than processing them sequentially. This allows the model to consider each word in the context of the entire sentence, not just the preceding words. Transformers form the foundation of several large language models, such as ChatGPT. Initially developed for natural language processing, they have since been adapted for tasks in the vision domain, such as the Vision Transformer (ViT) \cite{dosovitskiy2020image}.

However, in the vision domain, transformer-based models face several challenges, including high computational demands for image processing and the absence of inductive biases, such as translation invariance, which are inherent in traditional convolutional neural networks (CNNs). These models also typically require large training datasets and a significant number of parameters, making them computationally intensive and difficult to interpret.

\subsection{Top-down visual attention on convolutional neural networks}
\citeA{cao2015look} introduces a CNN that integrates top-down visual attention through feedback mechanisms. Unlike traditional feedforward CNNs, which process information in a bottom-up manner, the proposed model incorporates a feedback loop that adjusts the activation of hidden neurons based on high-level semantic information, such as class labels. This feedback is facilitated by ``gate layers’’, which learn gate values through the feedback update. These ``gate values’’ modify the outputs of the feedforward layers, allowing the network to focus on relevant image regions while suppressing irrelevant ones. Inspired by the top-down attention mechanisms of the human visual system, this method outperforms conventional models, such as GoogleNet and VGG, in weakly supervised object localization and image classification. The paper demonstrates that integrating feedback mechanisms into CNNs enhances selectivity, improves recognition accuracy, and provides a unified approach to both recognition and localization tasks. In our work, we also employ gate layers to introduce top-down attention and achieve selective processing.

\subsection{Studies of biological visual attention with machine learning models}
There are studies that study models of neural activities in visual attention with deep learning methods. For instance, \citeA{lindsay2018biological} used a deep learning model to simulate the feature similarity gain model of attention, where neural activity is enhanced for the target feature. By associating the tuning curve with the gradient, they found that the type and magnitude of neural modulations correspond to the type and magnitude of performance changes. They also discovered that attention applied at earlier stages has a weaker effect than at higher stages. Additionally, they identified an alternative method in which attention is applied according to values computed to optimally modulate higher areas and tested this possibility.

\section{Methods}
\subsection{Models}

\subsubsection{Dual-Network System}
In \citeA{pollack1987cascaded}, a connectionist network system was introduced, consisting of two networks: a function network, which combines inputs and produces outputs, and a context network, which takes a separate input and generates weights as output. These weights are then used by the function network to process its input. During training, the context network learns through ``cascaded back-propagation'', where the loss function is computed based on the ''function network’s output, and the loss is propagated through the function network to update the context network. The function network itself has no trainable parameters and relies entirely on the weights provided by the context network to determine its output.

Our dual-network system is a modification of Pollack’s system, incorporating both a context network and a function network. Unlike Pollack’s model, our function network has weights, which are learned during pretraining and then fixed.

Additionally, we introduce a gate layer in the function network, similar to the gate layer in \citeA{cao2015look}. These gates modulate the feature maps in the function network by pointwise multiplication, determining which features are passed forward and which are suppressed. This mechanism is essential for implementing both spatial and feature-based attention, allowing selective control over feature representation.

The gate values are computed by the context network, which learns optimal values based on the task via cascaded back-propagation. Errors are propagated through the function network and into the context network, enabling it to update the gates.

The dual-network system is designed to attend to a target object among multiple items and classify it. The input consists of an image containing multiple digits and a top-down signal indicating which digit to classify. For example, if the input image contains the digits 3 and 8, and the signal specifies 3 as the target, the correct classification output is 3.

The function network, which is pre-trained on a single-digit dataset, serves as a fixed processor. It can classify single digits but lacks the ability to handle multiple-digit classification on its own. The context network processes the top-down signal and directs the function network’s attention to the relevant digit by modulating the feature maps through gates.

This system implements two types of attention: spatial attention and feature-based attention. In the spatial attention setup, the context network receives a spatial signal (e.g., ``left'' or ``right'') and directs attention accordingly. In the feature-based attention setup, digits are divided into two categories: Group 1 (0–4) and Group 2 (5–9). Each input image contains one digit from each group, and the signal specifies whether to focus on Group 1 or Group 2. For example, if an image contains 6 and 0, and the signal specifies Group 1, the correct classification output is 0. If the signal specifies Group 2, the output is 6. Unlike spatial attention, which can rely solely on location-based cues, feature-based attention requires the context network to analyze the entire image to determine where the target digit is located.

\begin{figure}
\begin{center}
\includegraphics[scale=0.53]{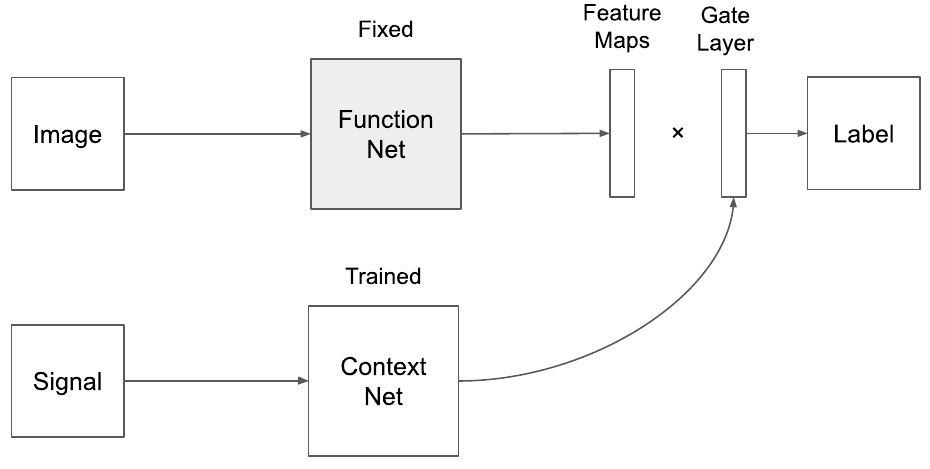}
\end{center}
\caption[Dual Network System for Spatial Attention]{Dual-network system for spatial attention. The context network only receives a top-down signal.}
\label{fig:model_spatial}
\end{figure}

\begin{figure}
\begin{center}
\includegraphics[scale=0.53]{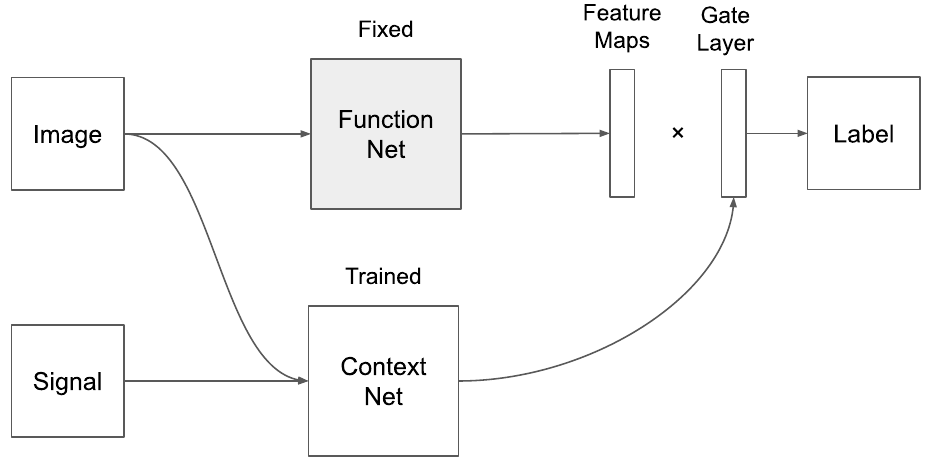}
\end{center}
\caption[Dual Network System for Feature-Based Attention]{Dual-network system for feature-based attention. The context network receives both the input image and a top-down signal.}
\label{fig:model_feature}
\end{figure}

Figures~\ref{fig:model_spatial} and~\ref{fig:model_feature} illustrate the architectures of the spatial attention and feature-based attention dual-network systems. The primary difference between these two architectures is that in feature-based attention, an additional connection from the input image to the context network is introduced. This connection allows the context network to analyze the content of the image rather than relying only on spatial location, so it can determine the location of the target digit.


\subsubsection{Function Network}
\begin{figure}
\begin{center}
\includegraphics[scale=0.55]{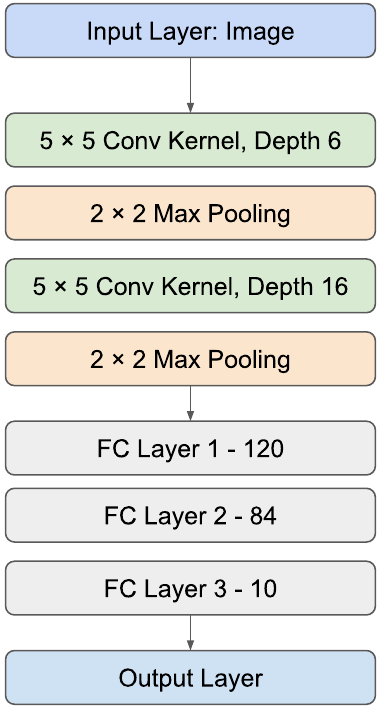}
\end{center}
\caption[Structure and Implementation Details of Function Network]{Structure and implementation details of function network.}
\label{fig:func_net}
\end{figure}

The function network is a CNN pre-trained on the MNIST dataset. Its implementation details are described in Figure~\ref{fig:func_net}. During training, the network achieved an average accuracy of 92\% on the single-digit classification task. The gates, generated by the context network, are applied at the second convolutional layer of the function network.

\subsubsection{Context Network}
The context network is trained to assist the function network in focusing on the target digit during the multi-digit task. Its training involves cascaded error back-propagation. The context network takes the top-down signal as input and generates gate values that match the size of the function network’s feature maps, which are then applied via pointwise multiplication.

\begin{figure}
\begin{center}
\includegraphics[scale=0.55]{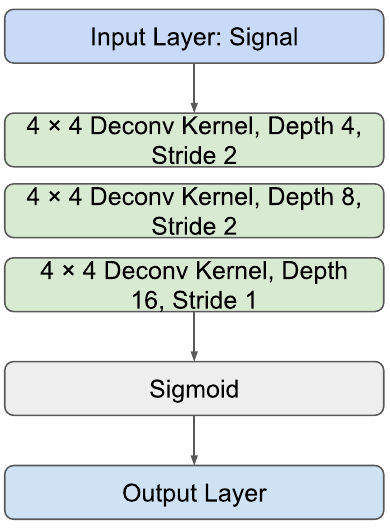}
\end{center}
\caption[Structure and Implementation Details of Context Network for Spatial Attention]{Structure and implementation details of context network for spatial attention}
\label{fig:conx_net_spatial}
\end{figure}

The context network has two versions. The first version, designed for spatial attention, does not process the input image. It only receives the spatial signal and generates gate values that correspond to the feature maps of the function network. The one-hot signal is expanded through deconvolutional layers to match the target feature map dimensions. The implementation details for this version are shown in Figure~\ref{fig:conx_net_spatial}.

\begin{figure}
\begin{center}
\includegraphics[scale=0.45]{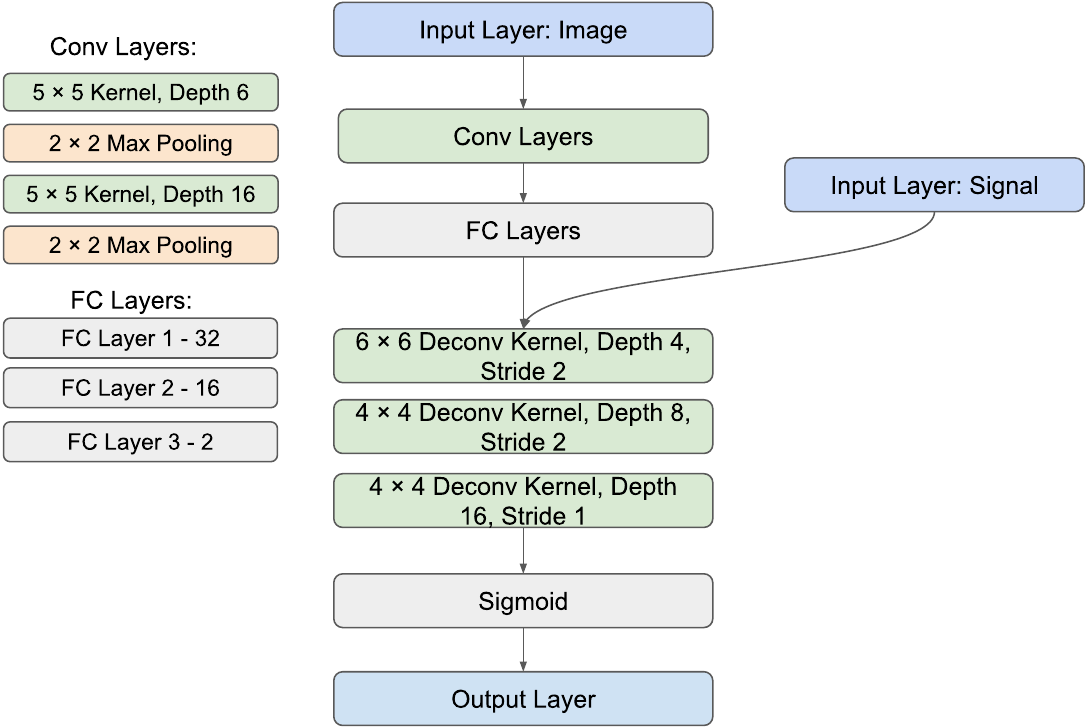}
\end{center}
\caption[Structure and Implementation Details of Context Network for Feature-Based Attention]{Structure and implementation details of context network for feature-based attention}
\label{fig:conx_net_feature}
\end{figure}

In the feature-based attention version, the context network requires access to the input image to determine which digit belongs to the target category. The input image, identical to the one used in the function network, is also fed into the context network. The context network extracts features using convolutional layers, concatenates them with the top-down signal, and then processes them through deconvolutional layers to generate gate values. Implementation details for the feature-based attention version are shown in Figure~\ref{fig:conx_net_feature}.

\subsection{Dataset}
\begin{figure}
\begin{center}
\begin{tabular}{ccc}
\includegraphics[scale=0.2]{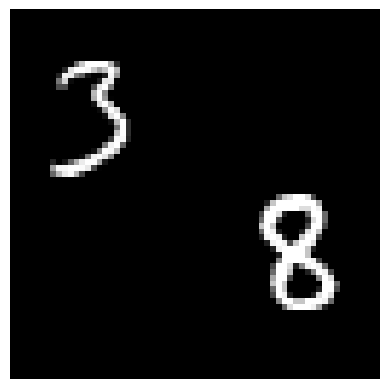}
& \includegraphics[scale=0.2]{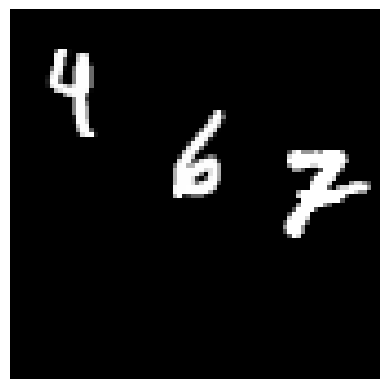}
& \includegraphics[scale=0.2]{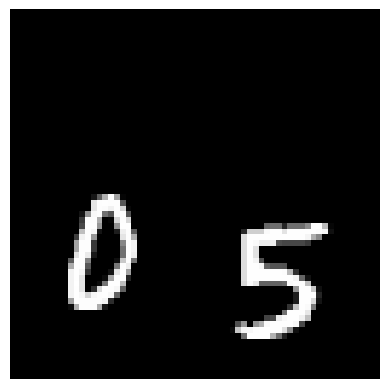}
\end{tabular}
\end{center}
\caption[Input Examples for Spatial and Feature-Based Attention Tasks]{Input examples for two-digit spatial attention task (left), three-digit spatial attention task (middle), and two-digit feature-based attention task (right). }
\label{fig:input_examples}
\end{figure}

We used a dataset named ``Multi-MNIST''\cite{mulitdigitmnist}. This dataset is based on the MNIST dataset \cite{deng2012mnist}. When training for feature-based attention, we make sure that two digits will always contain one digit from 0 to 4 and the other digit from 5 to 9. Input image examples are shown in Figure~\ref{fig:input_examples}.

\section{Results}
The mean classification accuracies for the three tasks—demonstrating spatial attention on a two-digit input image, spatial attention on a three-digit input image, and feature-based attention on a two-digit input image—are shown in Table~\ref{table:results}. For each task, the model was run five times to compute the mean classification accuracy and standard error.

Visualizations of the gate values generated by the dual-network system for the three tasks are shown in Figure~\ref{fig:vis_spatial_2}, Figure~\ref{fig:vis_spatial_3}, and Figure~\ref{fig:vis_feature_2}, respectively.

A Gaussian filter with a kernel size of (5, 5) and a sigma value of 0.8 was applied to the gate layers, followed by upsampling to match the input image size. The gate values range from 0 to 1, where brighter colors indicate values closer to 1 (attended), and darker colors represent values closer to 0 (not attended). For illustrative purposes, the gate values were superimposed onto the input images. In the experiment, the gate values were pointwise multiplied with the output from the second convolutional layer of the function network.

\subsection{Spatial attention on two digits}
First, we present the classification performance of the dual-network system on the two-digit dataset. The results are shown in the first data column in Table~\ref{table:results}.

The function network is pre-trained on the MNIST single-digit dataset. The current task uses a two-digit MNIST dataset and requires the network to classify one of the two digits according to a top-down signal. The top-down signal, in this case, is a spatial cue that indicates whether the classification target is the ``left'' or the ``right'' digit. The function network itself is unable to incorporate this additional input. With only the function network, the mean classification accuracy over five runs is 39.67 \%, with a standard error of 3.60E-04. Compared with the 92 \% accuracy on the single-digit MNIST dataset, this is a significant drop in performance.

In the dual-network system, the context network is responsible for processing the top-down signal and assisting the function network. After processing the signal, the context network generates a set of gate values to weigh the feature maps of the function network. With the help of the context network, the dual-network system achieves a mean classification accuracy of 93.62 \% with a standard error of 7.82E-04. This represents an improvement of 135.97 \% (3.41E-03) over the baseline accuracy.

\begin{figure}
\begin{center}
\begin{tabular}{c}
    {Input image} \\
    \includegraphics[scale=0.1]{vis_sa_2_1.png} \\
    {gate values}\\
    \includegraphics[scale=0.33]{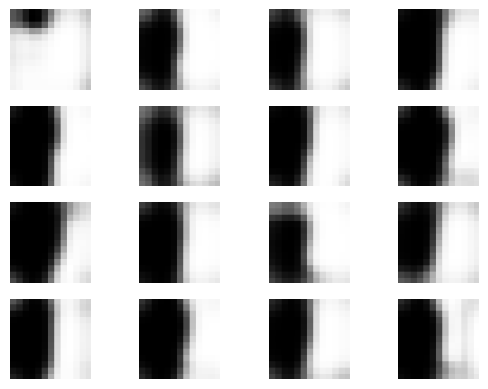} \\
    {gate values applied} \\
    \includegraphics[scale=0.33]{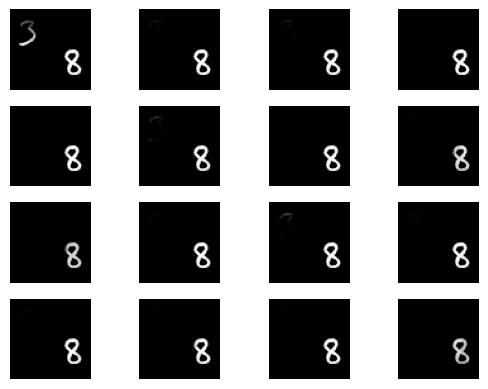} \\
\end{tabular}
\end{center}
\caption[Visualization of gate Values for Two Digit Spatial Attention Task]{The visualization of gate values for the two-digit spatial attention task. The top panel displays the input image. The middle panel displays the 16 channels of gate values. The bottom panel displays these channels superimposed on the input image.}
\label{fig:vis_spatial_2}
\end{figure}

A visualization of the gate layers is shown in Figure~\ref{fig:vis_spatial_2}. The gate layer consists of 16 channels, 15 of which show a clear division between the attended region (brighter areas, indicating gate values close to 1) and the unattended region (darker areas, indicating gate values close to 0). One channel is uniformly bright across the entire region. By superimposing the gate values onto the input image, we observe that the gate layer demonstrates spatial attention, allowing feature map values related to the target region to pass to the next stage, while filtering out irrelevant information in most channels (in the example illustrated in Figure~\ref{fig:vis_spatial_2}, the digit in the attended region is 8).

\subsection{Spatial attention on three digits}
The dual-network systems performed well in the previous task and demonstrated spatial attention. To test its generalizability, we increased the difficulty by using a three-digit MNIST dataset. The top-down signal in this task indicates either the ``left'', ``middle'', or ``right'' digit is the classification target. The second data column in Table~\ref{table:results} shows the mean classification accuracy of this task. Similarly, the pre-trained function had a baseline accuracy of 24.79 \% (9.02E-04) on the 3-digit dataset. With the context network, which receives the top-down signal and generates gate values, the accuracy improved to 89.44 \% (9.02E-04), a 260.85 \% (1.15E-02) increase.

\begin{figure}
\begin{center}
\begin{tabular}{c}
    {Input image} \\
    \includegraphics[scale=0.1]{vis_sa_3_1.png} \\
    {gate values}\\
    \includegraphics[scale=0.33]{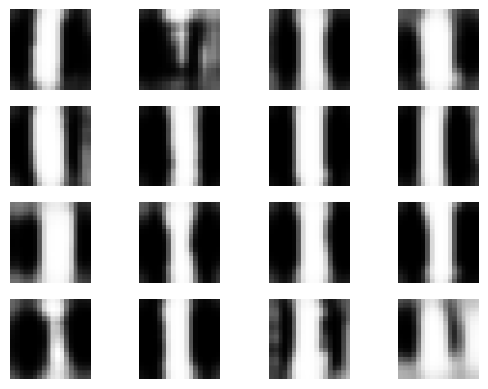} \\
    {gate values applied} \\
    \includegraphics[scale=0.33]{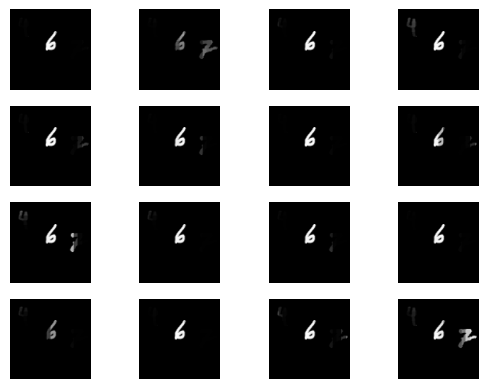} \\
\end{tabular}
\end{center}
\caption[Visualization of gate Values for Three Digits Spatial Attention Task]{The visualization of gate values for the three-digit spatial attention task. Similar format as in Figure~\ref{fig:vis_spatial_2}.}
\label{fig:vis_spatial_3}
\end{figure}

Figure~\ref{fig:vis_spatial_3} shows the visualization of the gate layers for the three-digit spatial top-down signal task. The gate layers learned to divide the image into three regions and highlight the region containing the target digit (in the example shown in Figure~\ref{fig:vis_spatial_3}, the middle region, where the target digit is 6). From the visualization of the gate values, we observe that almost all the channels assign high values (close to 1) to the target region. The dual-network system also demonstrated spatial attention in the three-digit dataset.

\subsection{Feature-based attention on two digits}
To demonstrate feature-based attention, we used a different top-down signal than in the previous two tasks. We first divided the 10 numerical digits into two groups: Group 1 consists of digits 0 to 4, and Group 2 consists of digits 5 to 9. This grouping is arbitrary, and any partitioning of digits 0 to 9 into two sets could be used for Group 1 and Group 2. The top-down signal in this task indicates the group of the target digit. Additionally, we modified the two-digit MNIST dataset so that each input image contains one digit from Group 1 and one digit from Group 2 (cases such as 0 and 3, where both digits belong to the same group, were excluded).

For this task, the model learned more slowly, so we increased the dataset size to 160000 training images and 40000 testing images. In each run, the model was trained for 10 epochs. In contrast, for the two previous tasks, both datasets contained 80000 training images and 20000 testing images, and the models were trained for 5 epochs.

Classification results are shown in the third data column of Table~\ref{table:results}. The baseline accuracy is 34.96 \%, with a standard error of 1.53E-03. The dual-network system achieves a mean classification accuracy of 90.52 \%, with a standard error of 1.05E-02, representing an improvement of 155.88 \% (2.22E-02).

\begin{figure}
\begin{center}
\begin{tabular}{c}
    {Input image} \\
    \includegraphics[scale=0.1]{vis_FBA_2_1.png} \\
    {gate values} \\
    \includegraphics[scale=0.33]{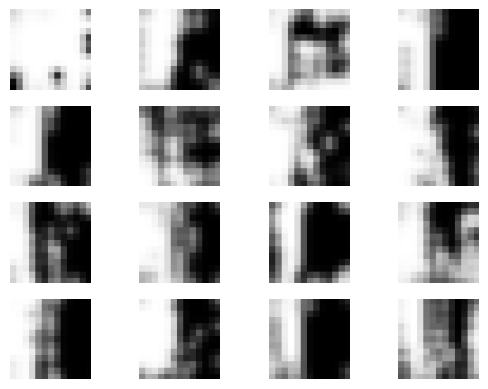} \\
    {gate values applied} \\
    \includegraphics[scale=0.33]{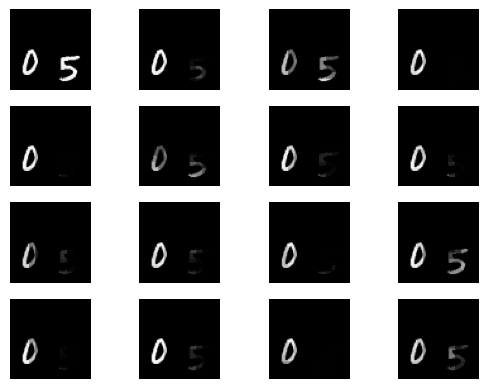} \\
\end{tabular}
\end{center}
\caption[Visualization of Gate Values for Two Digit Feature-Based Attention Task]{The visualization of gate values for the two-digit feature-based attention task. Similar format as in Figure~\ref{fig:vis_spatial_2}.}
\label{fig:vis_feature_2}
\end{figure}

The visualization is shown in Figure~\ref{fig:vis_feature_2}. Gate values for feature-based attention differ from those for spatial attention, appearing more chaotic. However, they still demonstrate an attention effect, where the non-target digit is not attended to.

\begin{table*}
\caption[Mean Classification Accuracies with Standard Errors for Three Tasks]{Mean classification accuracies (\%) for three tasks. The values in parentheses indicate standard errors. The classification accuracies are averaged over 5 runs for each task.}
\label{table:results}
\centering
\begin{tabular}{ccccl}
\multirow{2}{*}{Model} & \multicolumn{4}{c}{Tasks} \\ \cline{2-5} 
 & Spatial (2 digits) & Spatial (3 digits) & \multicolumn{2}{c}{Feature (2 digits)} \\ \cline{2-5} 
FN & 39.67\% (3.60E-04) & 24.79\% (9.02E-04) & \multicolumn{2}{c}{34.96\% (1.53E-03)} \\
FN + CN & 93.62\% (7.82E-04) & 89.44\% (1.02E-03) & \multicolumn{2}{c}{90.52\% (1.05E-02)} \\
Improvement & 135.97\% (3.41E-03) & 260.85\% (1.15E-02) & \multicolumn{2}{c}{158.88\% (2.22E-02)} \\ \hline
\end{tabular}
\end{table*}

\section{Discussion}

In this work, we presented a dual-network system to simulate top-down control in visual attention. The system comprises a fixed, pre-trained function network for single-digit classification and a trainable context network that uses top-down signals to generate gate values. These gate values prioritize task-relevant information, enabling the function network to adapt to multi-digit tasks. This approach effectively models spatial and feature-based attention, demonstrating how top-down signals enhance task performance.

Our findings indicate that top-down control of attention is spontaneous and learnable. The context network effectively learned to apply spatial or feature-based attention based on the training data rather than relying on hard-coded instructions to gate values out the left or right portions of images. This adaptability shows how top-down control of attention enables a simpler, single-task processor to handle more complex tasks.

The results of this study align with previous research, particularly in demonstrating that top-down attention mechanisms enhance task performance by guiding the selection of relevant information. This reinforces the notion that attention—whether in biological systems or neural network models—plays a crucial role in enhancing cognitive processing and performance in visual tasks. The consistency of our findings with existing literature further supports the idea that attentional guidance is a fundamental aspect of cognitive function.

One limitation of our study is the simplicity of the dataset, which contains only digits. Future research could benefit from utilizing more realistic image datasets, such as CIFAR-100, which includes multiple category structures, or MS-COCO, which contains multiple objects in a single image. These datasets would allow for a more comprehensive exploration of visual attention mechanisms in realistic and complex visual environments.

In conclusion, this work studies the critical role of top-down attention mechanisms in visual processing. By demonstrating that the dual-network system can significantly improve task performance, our findings provide valuable insights into how top-down contextual information helps to prioritize task-relevant input information and improve task performance. This work advances our understanding of top-down attention on visual perception and establishes a connection between computer and biological models of visual attention.

\clearpage

\bibliographystyle{apacite}

\setlength{\bibleftmargin}{.125in}
\setlength{\bibindent}{-\bibleftmargin}

\bibliography{reference}

\typeout{get arXiv to do 4 passes: Label(s) may have changed. Rerun}
\end{document}